\documentclass{article}
\usepackage[letterpaper,textwidth=5.5in,textheight=9in,top=1in,footskip=30pt]{geometry}
\usepackage{times}
\usepackage{natbib}
\usepackage{fancyhdr}
\makeatletter
\renewcommand{\normalsize}{%
  \@setfontsize\normalsize\@xpt\@xipt
  \abovedisplayskip 7\p@ \@plus 2\p@ \@minus 5\p@
  \abovedisplayshortskip \z@ \@plus 3\p@
  \belowdisplayskip \abovedisplayskip
  \belowdisplayshortskip 4\p@ \@plus 3\p@ \@minus 3\p@
}
\normalsize
\renewcommand{\section}{%
  \@startsection{section}{1}{\z@}
    {-2.0ex \@plus -0.5ex \@minus -0.2ex}
    {1.5ex \@plus 0.3ex \@minus 0.2ex}
    {\large\bfseries\raggedright}}
\renewcommand{\subsection}{%
  \@startsection{subsection}{2}{\z@}
    {-1.8ex \@plus -0.5ex \@minus -0.2ex}
    {0.8ex \@plus 0.2ex}
    {\normalsize\bfseries\raggedright}}
\renewcommand{\maketitle}{%
  \thispagestyle{preprintfirst}
  \begin{center}
    \hrule height 4pt
    \vspace{0.25in}
    {\LARGE\bfseries\@title\par}
    \vspace{0.29in}
    \hrule height 1pt
    \vspace{0.22in}
    {\bfseries\@author\par}
  \end{center}
  \vspace{0.2in}
}
\makeatother
\fancypagestyle{preprintfirst}{%
  \fancyhf{}
  \fancyfoot[L]{\footnotesize Preprint.}
  
}
\renewenvironment{abstract}
  {\begin{center}\large\bfseries Abstract\end{center}\begin{quote}}
  {\end{quote}\vspace{1ex}}
\usepackage[utf8]{inputenc}
\usepackage[T1]{fontenc}
\usepackage{hyperref}
\hypersetup{hidelinks}
\usepackage{url}
\usepackage{booktabs}
\usepackage{amsmath,amsfonts,amssymb}
\usepackage{microtype}
\usepackage{xcolor}
\usepackage{graphicx}
\usepackage{subcaption}
\usepackage{float}
\usepackage{enumitem}
\usepackage{multirow}
\newcommand{\persist}{\textsc{Persist}}
\newcommand{\replan}{\textsc{Replan}}

\title{A Score Is Not a Policy: Measuring the Value of Adaptive Revision}
\author{Ayushi Chadha\\
\texttt{ayushichadha48@gmail.com}}

\begin{document}
\maketitle

\begin{abstract}
As agentic systems become compound systems, increasingly important decisions move above task execution itself: when should a higher-level controller preserve the strategy guiding another process, and when should it revise it? We study this meta-level control problem in a hierarchical latent reasoner whose manager can retain or replace a commitment governing lower-level computation. Across three precommitted training seeds, learned revision timing produces qualitatively different policies, ranging from an almost deterministic early clock to substantially more state-conditioned schedule distributions, yet none outperforms the best forced timing policy evaluated on the same frozen checkpoint. This separates state dependence from decision value: a controller can vary its actions with internal state without turning that variation into a reproducible task-performance benefit. A deeper intervention study on the original checkpoint shows that timing itself is consequential and order-sensitive, while exhaustive enumeration reveals that a strong fixed schedule captures most of the measurable value available from timing at this decision budget. Counterfactual \persist/\replan{} diagnostics further show why score-level evidence can be misleading when predictability is dominated by decision position rather than within-position discrimination. Together, these results argue that learned meta-level control should be evaluated along three separate axes: whether its score depends on state, whether that dependence changes realized behavior, and whether those changes capture outcome value beyond a strong non-adaptive policy.
\end{abstract}

\section{Introduction}

Compound AI systems increasingly contain decisions that sit above ordinary task execution. A higher-level process may decide whether to preserve or revise a prompt, trajectory, program, subgoal, memory state, or agent harness. Recent meta-level optimizers make this structure explicit: GEPA learns from system trajectories to revise prompts \citep{agrawal2025gepa}, AlphaEvolve repeatedly proposes and evaluates program modifications \citep{alphaevolve2025}, and Meta-Harness searches over executable harnesses using scores and traces from prior candidates \citep{lee2026metaharness}. These systems differ substantially, but they share a basic problem: a meta-level decision must be assigned credit for its downstream consequences. The quality of the score used to train or rank that decision is therefore not merely an implementation detail.

This paper isolates that problem in a setting where the relevant decisions can be measured directly. We use a compact hierarchical latent reasoner derived from the Hierarchical Reasoning Model (HRM) \citep{wang2025hrm}. A slow manager emits a directional latent commitment that biases a faster worker for multiple refinement passes. The meta-level action is deliberately simple: after observing the current pass, should the manager \emph{retain} the active commitment or \emph{replace} it for subsequent computation? This turns subgoal persistence into a small supervisory-control problem with a fixed compute budget and an exactly enumerable timing space.

The study began with a natural hypothesis: a state-conditioned trigger should allocate a limited revision budget more intelligently than a fixed clock. A first 400-step run instead collapsed almost completely to the earliest eligible second intervention. We then preregistered two additional adaptive seeds under the same training, budget-calibration, and evaluation protocol and retained all outcomes regardless of direction. They did not reproduce the same clock: one produced a high-entropy mixture of schedules and the other mostly retained the initial commitment. What did replicate was more consequential: in no seed did the learned adaptive policy outperform the best forced timing policy on the same frozen checkpoint. Thus the replicated failure is not universal clock collapse, but a lack of reproducible adaptive value despite qualitatively different degrees of state-conditioned behavior.

This distinction motivates the central claim of the paper: \textbf{state dependence is not adaptation, and adaptation is not yet utility}. A score may vary with internal state without changing actions; actions may vary substantially across episodes without improving outcomes; and even a genuinely state-aware policy may have little value if a strong fixed decision already captures most of the available gain. These possibilities are easy to conflate when only the controller's score or schedule diversity is inspected.

Our contributions are fourfold. First, we provide a three-seed, precommitted matched-budget replication of learned revision timing and show that policy form is seed-sensitive while controlled adaptive advantage is absent in all three runs. Second, we use frozen-checkpoint interventions to localize the seed-0 failure to timing rather than generally weak representations. Third, we exhaustively enumerate the complete two-intervention timing space on that checkpoint and quantify both order effects and the oracle ceiling on episode-specific adaptation. Fourth, we use matched counterfactual \persist/\replan{} branches to separate cross-position predictability from within-position decision signal. The resulting protocol, comprising replication, behavioral audit, frozen intervention, exhaustive decision-value measurement, and counterfactual branching, is intended as a diagnostic template for meta-level control, not as a claim that this small internal manager is itself a full meta-agent.

\section{Supervisory Replanning in a Hierarchical Reasoner}
\label{sec:setting}

We study supervisory revision in a compact two-level latent reasoner built on HRM \citep{wang2025hrm}. The architecture is used as a controlled testbed rather than as a claim about the necessity of hierarchy: it exposes a persistent, addressable latent commitment whose timing can be intervened on exactly. A slow \emph{manager} proposes the commitment; a fast \emph{worker} performs the task-conditioned refinement that consumes it.

\subsection{Manager--worker commitments}
Let $m\in\{1,\ldots,M\}$ index outer refinement passes. The model maintains high- and low-level hidden sequences $z^H_m,z^L_m$. From the post-pass high-level representation $h_m=z^H_m[:,0]$, the manager produces a normalized directional goal and a scalar gate,
\begin{align}
\tilde g_m &= s_g\frac{W_g h_m+b_g}{\max(\lVert W_g h_m+b_g\rVert_2,\epsilon)},\\
\alpha_m &= \sigma\!\left((w_\alpha^\top h_m+b_\alpha)/\tau\right).
\end{align}
The goal is projected back into the shared hidden space and injected into subsequent low- and high-level updates with strength $\alpha_m$. The pooled worker representation is $w_m=T^{-1}\sum_t z^L_{m,t}$. When a commitment is emitted after pass $s$, the contemporaneous worker representation is stored as an anchor $a=w_s$. On later passes consuming that commitment, the directional auxiliary objective measures displacement relative to the anchor,
\begin{equation}
\ell^F_m=A_m\alpha_m\left[1-\cos(w_m-a_m,g_m)\right],
\end{equation}
where $A_m$ masks passes on which no causal commitment has yet been consumed. The goal, gate, and anchor persist until the next manager intervention. The \emph{dwell} of a commitment is therefore the number of worker passes for which this tuple remains active. This construction follows the directional-control lineage of FeUdal Networks \citep{vezhnevets2017fun}, while our question concerns persistence and revision rather than reward-maximizing hierarchical control.

\subsection{Causal intervention timing}
All primary experiments execute a fixed $M=8$ outer passes. At pass $m$, the worker first consumes the commitment already stored in recurrent carry; the hidden states are updated; only then does the manager observe the post-pass state and decide whether to retain or replace the commitment. A replacement decided after pass $m$ is first consumed on pass $m+1$ and can never affect pass $m$ itself. Pass 1 is a bootstrap pass followed by a mandatory initial emission. Optional revision is eligible after passes 2--7, and no emission is allowed after pass 8 because no subsequent worker computation could consume it.

\begin{figure}[t]
  \centering
  \includegraphics[width=0.88\linewidth]{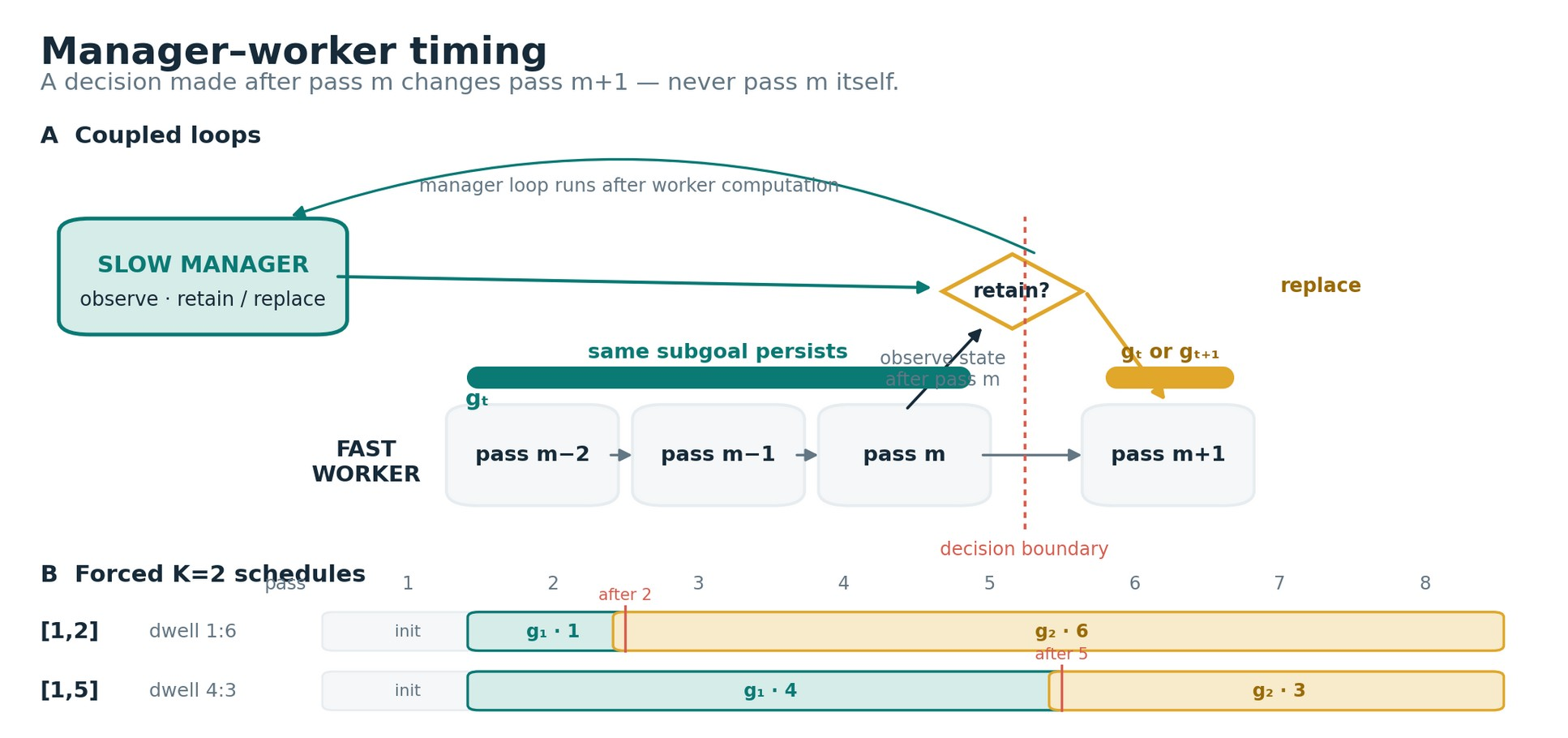}
  \caption{Manager--worker timing. A decision made after pass $m$ changes the commitment first consumed on pass $m+1$. Bottom: $[1,2]$ and $[1,5]$ use the same $K=2$ intervention budget but allocate persistence differently.}
  \label{fig:arch}
\end{figure}

We write a schedule as the ordered list of emission positions. Thus $[1,4]$ denotes the mandatory emission after pass 1 followed by one replacement after pass 4, yielding commitment dwells of 3 and 4 worker passes. Under $M=8$ and exactly two emissions ($K=2$), the complete fixed-clock space is $[1,k]$ for $k\in\{2,3,4,5,6,7\}$. Figure~\ref{fig:arch} makes the emission/consumption boundary explicit.

\subsection{From a clock to a state-conditioned trigger}
A fixed-period manager implements a temporal policy $u_m=f(m)$, where $u_m\in\{0,1\}$ denotes replacement after pass $m$. The adaptive manager instead observes a compact detached state summary
\begin{equation}
\phi_m=[c_m,d_m,\rho_m,\tau_m/M,q_m,\alpha_m]^\top,
\end{equation}
where $c_m=\cos(w_m-a_m,g_m)$ is cumulative directional progress, $d_m=\cos(w_m-w_{m-1},g_m)$ is latest-pass progress, $\rho_m=c_m-c_{m-1}$ is its trend, $\tau_m/M$ is normalized dwell, $q_m$ is a bounded halt-versus-continue confidence signal, and $\alpha_m$ is the active gate. All six features are detached before entering the trigger. An affine Bernoulli policy produces
\begin{equation}
\beta_m=\sigma(w_\beta^\top\phi_m+b_\beta),\qquad u_m\sim\mathrm{Bernoulli}(\beta_m)
\end{equation}
during training, while deterministic evaluation uses $u_m=\mathbb{1}[\beta_m>\theta]$. The six-feature trigger is a project-specific synthesis: options and Option-Critic motivate persistent behavior with state-dependent termination \citep{sutton1999options,bacon2017optioncritic}, while ACT, PonderNet, and Skip-RNN motivate learned continuation decisions \citep{graves2016act,banino2021pondernet,campos2018skiprnn}. Here the persistent object is a latent directional commitment inside iterative reasoning.

\subsection{One-consuming-pass credit}
The causal boundary also constrains training. A decision after pass $m$ first changes pass $m+1$, but the recurrent training loop performs backward propagation and an optimizer update between outer refinement calls. We therefore do not carry an unrestricted graph through the full future trajectory. Cached commitments are stored as detached values and, on the first consuming pass, a local value-preserving Jacobian is reconstructed:
\begin{equation}
v_{\mathrm{used}}=v_{\mathrm{cached}}+I_{\mathrm{pending}}\left(v_{\mathrm{recomputed}}-\operatorname{stopgrad}(v_{\mathrm{recomputed}})\right).
\end{equation}
The added term is numerically zero in the forward pass, so the worker consumes exactly the cached commitment; backward propagation follows the locally recomputed quantity. The construction is applied to goal, gate, and adaptive retain/replace selection. Thus the intervention receives credit from the \emph{first worker pass that consumes it}, not from unrestricted backpropagation through the complete refinement trajectory. This implementation is causally correct for the intended local credit path, but whether that horizon values revision in a way aligned with final task performance is an empirical question addressed in Section~\ref{sec:audit}.

\section{Experimental Protocol and Controlled Evaluation}
\label{sec:protocol}

Our design separates three quantities that are easily confounded in learned supervisory control: reasoner quality, intervention budget, and intervention timing. We first establish fixed-clock controls, then train an adaptive controller to the same optimization budget, calibrate its realized intervention count without using task performance, and finally freeze model parameters while intervening directly on timing.

\subsection{Task, compute, and evidence hierarchy}
All primary experiments use ConceptARC-mini with $M=8$ outer passes. Fixing $M$ ensures that changing a schedule changes \emph{when} commitments are replaced rather than how much outer reasoning compute is available. The corrected fixed-period baseline contains 15 deterministic training runs, $P\in\{1,3,4,6,\infty\}$ and seeds $\{0,1,2\}$, each trained for 400 optimizer steps. The adaptive controller is trained for 400 steps under three precommitted seeds. Seed 0 is the checkpoint for which we additionally performed the exhaustive six-clock sweep, oracle analysis, and counterfactual diagnostics; seeds 1 and 2 are independent replications of the matched-budget adaptive audit and frozen forced-schedule comparison.

The primary evidence is the 400-step matched-budget adaptive audit, the frozen-checkpoint schedule intervention, and the preregistered exhaustive $K=2$ sweep. Earlier 96-step adaptive runs establish mechanism sanity, including gradient flow, deterministic score ordering, causal retain/replace execution, and threshold behavior, but are not used for adaptive-versus-fixed performance claims. The later counterfactual-v2 experiment is a diagnostic and stops before training a new policy.

\subsection{Why \texorpdfstring{$K=2$}{K=2} is the controlled budget}
Across the three-seed fixed-period matrix, differences between persistence periods are small relative to across-seed variation, so the fixed sweep does not identify a universally optimal period. It does, however, contain a useful matched-count contrast: $P=4$ and $P=6$ both emit exactly $K=2$ commitments per episode but place the second emission at different positions. $P=6$ exceeds $P=4$ on all three seeds, with a mean held-out token-accuracy difference of 0.00248. The effect is modest, but it isolates timing from intervention count and motivates asking whether the second intervention can be allocated from state rather than a clock.

\subsection{Leakage-free budget calibration}
The adaptive controller is trained with no explicit intervention penalty ($\eta=0$). During the 400-step run, stochastic training averages 3.945 total emissions per completed episode. At deterministic evaluation with the nominal $\theta=0.5$, it retains after the mandatory initial commitment. Because the learned score is reproducibly non-constant, we treat this as an operating-point problem rather than evidence that the trigger is dead.

For matched evaluation, $\theta$ is calibrated on a split disjoint from final evaluation using \emph{intervention count only},
\begin{equation}
\theta^*=\arg\min_\theta |\bar K_{\mathrm{cal}}(\theta)-2|.
\end{equation}
No accuracy or loss enters threshold selection. The chosen $\theta^*=0.498200$ gives $\bar K_{\mathrm{cal}}=2.000000$ and, once frozen, $\bar K_{\mathrm{final}}=2.000814$ on the ordered 3,686-episode final split. We therefore match the decision budget without treating $\beta_m$ as a calibrated probability of replanning benefit.

\subsection{Frozen timing interventions and exhaustive enumeration}
The strongest control freezes the exact adaptive step-400 checkpoint and overrides only the position of its second commitment. Under forced $[1,k]$, the model follows the normal causal trajectory and uses the manager-generated goal at the forced emission state; no parameter is retrained. This holds fixed the learned substrate and changes only the supervisory timing rule.

Because $M=8$ and $K=2$, all six legal schedules can then be evaluated on the same frozen checkpoint and the same ordered episodes. Before running this sweep, we froze the checkpoint, final split, compute semantics, six schedules, headline micro-token metric, and reversed-dwell hypotheses. Schedule and oracle comparisons use 10,000 paired episode-bootstrap resamples. Micro token accuracy is the preregistered headline for the sweep; earlier matched-controller tables use episode-averaged token accuracy and are labeled separately. Exact sequence accuracy is zero in this compact regime and is not used for schedule comparison.

The original fixed $P=4$ and $P=6$ runs did not persist final weights. We deterministically reran their recorded seed-0 commands; the resulting training summaries matched the original records to reported precision. The matched-controller values reported here, 0.481321 and 0.485595, come from re-evaluating those reconstructed checkpoints on the ordered 3,686-episode final split. All causal frozen-schedule claims instead use the explicitly saved and hashed adaptive checkpoint.

\section{Learned Replanning Is State-Dependent but Not Reliably Better}
\label{sec:collapse}

\subsection{Three precommitted seeds produce different policies}
The three adaptive runs reach final episode-averaged token accuracies of 0.467015, 0.486149, and 0.474947, with held-out intervention counts $K=2.0008$, $1.9674$, and $1.9135$. Table~\ref{tab:replication} isolates the controlled performance comparison; the score-structure diagnostics below characterize how each learned policy realizes that budget. Their realized policies differ qualitatively. Seed 0 is almost deterministic: 99.19\% of episodes use $[1,2]$, schedule entropy is 0.0776 bits, position explains 96.76\% of $\beta$ variance, and a time-only score has Spearman $\rho=0.9776$ with the full trigger. Seed 1 instead has 2.8598 bits of schedule entropy, only 60.77\% position-explained variance, and time-only/full-score $\rho=-0.2799$. Seed 2 mostly emits only the mandatory initial commitment ($[1]$ on 78.59\% of episodes), but its score is likewise not clock-like under the same diagnostic (54.51\% position-explained variance; $\rho=-0.2457$). Thus the exact clock-collapse phenomenon is specific to seed 0; learned policy form is not stable across training seeds.

\begin{table}[H]
\centering
\caption{\textbf{Adaptive timing does not outperform the best forced schedule across seeds.} Accuracy is episode-averaged token accuracy on the same frozen checkpoint; gap is adaptive minus best forced, in percentage points.}
\label{tab:replication}
\small
\setlength{\tabcolsep}{4.2pt}
\begin{tabular}{@{}crrrc@{}}
\toprule
Seed & Adaptive & Best forced & Gap & Dominant policy \\
 & acc. (\%) & acc. (\%) & (pp) & \\
\midrule
0 & 46.70 & $[1,4]$\;48.75 & $-2.05$ & $[1,2]$ (99.2\%) \\
1 & 48.61 & $[1,6]$\;48.95 & $-0.34$ & $[1]$ (34.2\%) \\
2 & 47.49 & $[1,2]$\;47.55 & $-0.06$ & $[1]$ (78.6\%) \\
\bottomrule
\end{tabular}
\end{table}

\subsection{State-conditioned behavior does not establish useful adaptation}
On every frozen checkpoint, the adaptive policy falls below the best tested forced two-update schedule: by 2.0496, 0.3386, and 0.0568 percentage points for seeds 0--2, respectively (Table~\ref{tab:replication}). Mean adaptive accuracy is 47.6037\%, versus 48.4187\% for the best within-checkpoint forced schedule. We do not treat this mean as a population estimate from three seeds; its role is descriptive. More importantly, seed 1 rules out the simplest interpretation of the original failure. Its controller is behaviorally diverse and substantially less position-dominated, yet that diversity does not produce a controlled accuracy advantage. Seed 2 is approximately neutral but likewise provides no positive evidence of useful timing. The replicated result is therefore narrower and stronger than ``the controller becomes a clock'': state-conditioned schedules emerge, but useful state-conditioned timing does not replicate.

\subsection{Seed 0 localizes a timing failure under frozen intervention}

A cross-checkpoint comparison cannot determine whether the adaptive model simply learned representations for which early replacement is appropriate. We therefore force alternative schedules on the same adaptive weights. Forced $[1,2]$ reproduces the calibrated adaptive result (0.467002 episode-averaged token accuracy). Delaying the second emission to $[1,4]$ or $[1,6]$ raises accuracy to 0.487511 and 0.486622 respectively, with no retraining and the same $K=2$ budget. The checkpoint therefore contains manager and worker representations capable of better performance under later intervention. The immediate failure is the controller's choice of \emph{when} to replace the commitment.

\subsection{The complete timing landscape is narrow but order-sensitive}
\begin{figure*}[t]
\centering
\begin{subfigure}[t]{0.44\textwidth}
\centering\includegraphics[width=\linewidth]{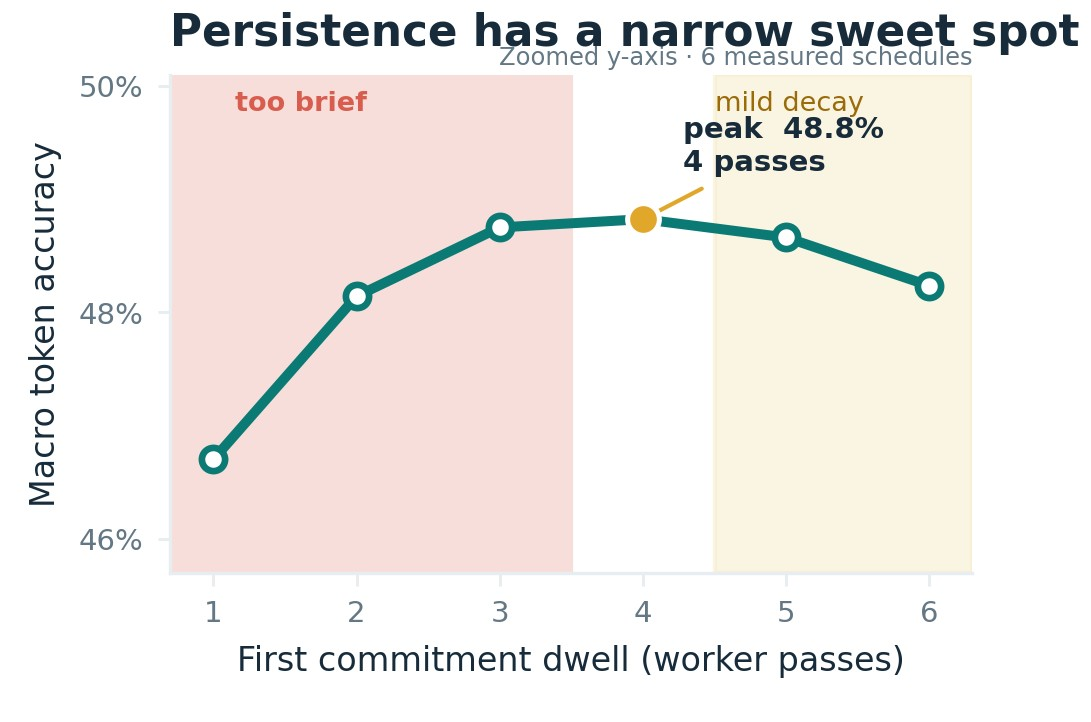}
\caption{Complete $K=2$ timing sweep.}
\end{subfigure}\hfill
\begin{subfigure}[t]{0.44\textwidth}
\centering\includegraphics[width=\linewidth]{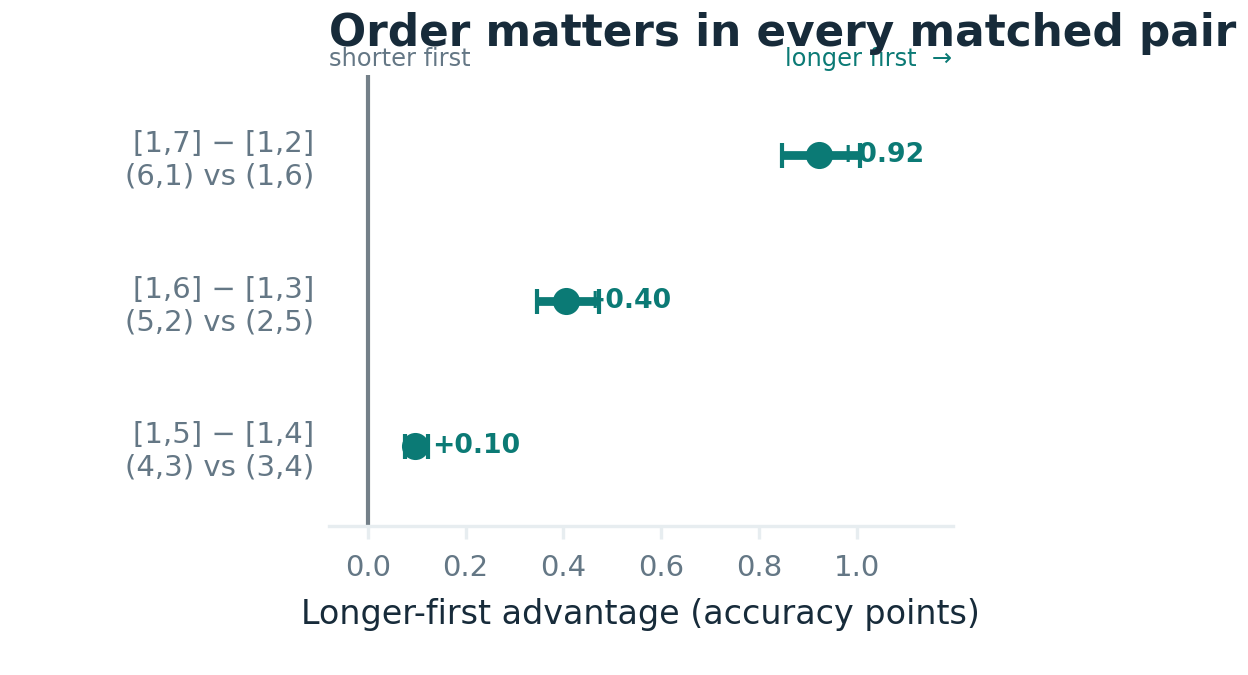}
\caption{Preregistered reversed-order tests.}
\end{subfigure}
\caption{Timing is consequential and order-sensitive. Left: performance has a narrow persistence sweet spot later than the learned controller's near-universal $[1,2]$ replacement. Right: placing the longer dwell first wins all three reversed pairs; intervals are paired episode-bootstrap 95\% CIs.}
\label{fig:timing}
\end{figure*}

The exhaustive sweep evaluates $[1,k]$ for every $k\in\{2,\ldots,7\}$ on the frozen checkpoint. Micro token accuracy rises sharply as the first commitment is allowed to persist, peaks at $[1,5]$, and declines gently thereafter (Figure~\ref{fig:timing}). The best observed fixed clock reaches 0.521206 micro accuracy. The numerical identity of $[1,5]$ is checkpoint-specific; the important fact is that the learned policy concentrates on an early region that the controlled endpoint measurement identifies as poor.

With seven consuming passes and $K=2$, reversed schedules exchange the same pair of dwell lengths: $[1,2]\leftrightarrow[1,7]$, $[1,3]\leftrightarrow[1,6]$, and $[1,4]\leftrightarrow[1,5]$. Before evaluation we predicted equality within each pair if only the unordered dwell multiset mattered. All three predictions are rejected. Expressed as longer-first minus shorter-first micro accuracy, the differences are 0.00924, 0.00405, and 0.00097, with paired-bootstrap 95\% intervals excluding zero in every case (Figure~\ref{fig:timing}). Thus persistence duration alone does not determine value: where a duration occurs in the reasoning trajectory matters. Because the two dwells sum to a constant, this design does not separately identify causal first- and second-dwell effects; it establishes order sensitivity within the tested anti-diagonal.

Across seeds, the learned policy does not recover a reproducible timing advantage, even though its behavioral form ranges from nearly deterministic to highly diverse. Seed 0 then provides a controlled mechanistic case study: forcing later timing on the same weights recovers performance, and exhaustive intervention locates its learned schedule on the wrong side of a non-monotonic timing landscape. We next ask why score-level evidence can still look informative without establishing useful adaptation.

\section{Auditing the Supervisory Signal}
\label{sec:audit}

\subsection{State-dependent score variation is not yet adaptive behavior}
An early 96-step checkpoint is useful as a mechanism audit because its score remains dispersed enough to inspect without retraining. At the natural $\theta=0.5$, all optional decisions are retained, yet a threshold sweep reveals 16.724\% within-position score variance and, at intermediate operating points, up to 21 distinct schedules with 2.344 bits of schedule entropy. Repeated extraction reproduces the ranking exactly and higher-precision recomputation produces no strict reversals between distinct float32 score levels. The trigger therefore learned a stable relative ordering even though the default threshold yields retain-all behavior.

The step-400 checkpoint sharpens the distinction: score variation remains, but temporal structure now explains 96.760\% of it and the calibrated policy executes $[1,2]$ almost universally. Hence
\begin{equation}
\text{state-dependent score}\;\not\Rightarrow\;\text{state-dependent realized policy}.
\end{equation}
The relevant audit must therefore ask whether states at the \emph{same decision position} are ranked usefully, not merely whether scores differ across the trajectory.

\subsection{Matched counterfactual branching}
We freeze the step-400 checkpoint and construct a local counterfactual target at every eligible state. From the identical detached latent carry, one branch keeps the old commitment and the other installs the candidate commitment. Each branch uses the selected commitment for one consuming transition followed by one Q-lookahead transition. For branch $b$, let
\begin{equation}
J_b=\ell_{\mathrm{LM},b}+\tfrac12(\ell_{\mathrm{halt},b}+\ell_{\mathrm{continue},b})+\lambda_F\ell_{F,b},
\end{equation}
and define detached replanning advantage
\begin{equation}
A=J_{\mathrm{persist}}-J_{\mathrm{replan}},
\end{equation}
so $A>0$ favors replacement. This is more causal than reading the learned trigger score because the two branches begin from the same state and differ only in commitment choice. It remains a bounded local surrogate, not full-episode return.

Across 64 episodes and 384 eligible states, PERSIST wins 90.365\% of branches. Mean and median $A$ are negative. A tiny critic using only cumulative progress and old-versus-candidate goal disagreement nevertheless predicts the target well in aggregate: held-out $R^2=0.406$, Spearman $\rho=0.729$, and sign accuracy 95.6\%. Taken alone, these metrics would suggest a strong state signal.

\subsection{Conditioning on position changes the conclusion}
Pass position explains 79.464\% of the critic's prediction variance. A pass-only predictor already achieves reconstructed held-out $R^2=0.292$. After removing training-split pass means from features and target, adding the residual critic improves full-advantage prediction, but held-out $R^2$ on the \emph{residual target itself} is $-0.007$.

Classification exposes the same issue more starkly. Of 114 held-out decision states, only 12 favor REPLAN, and all 12 occur at pass 2. A pass-only predictor has no ability to rank states within a pass, yet achieves pooled AUROC 0.966. High aggregate AUROC can therefore be obtained by learning \emph{where in time} favorable decisions occur while assigning identical scores to all states facing the same decision. Adding residual state information improves rare-state balanced accuracy, so state information is not absent; the evidence is simply too sparse and position-concentrated to justify a new end-to-end controller.

\subsection{Credit horizon versus endpoint value}
The local target and endpoint sweep measure different quantities, but their disagreement is precisely the design warning. The local diagnostic concentrates its positive replanning cases at the earliest eligible position, while final-outcome intervention shows that repeatedly selecting this early position is poor and that later fixed timing is substantially better on the same checkpoint. We therefore do not claim a pointwise theorem that every locally favorable revision is globally harmful. The supported conclusion is narrower and more useful:
\begin{equation}
\text{locally predictive revision signal}\;\not\Rightarrow\;\text{correct endpoint timing policy}.
\end{equation}
Implementation correctness guarantees that one-consuming-pass credit reaches the trigger; it does not guarantee that this horizon values intervention on the horizon that ultimately matters. The diagnostic was therefore used as a stop rule: globally predictive but weak within-position evidence was not treated as sufficient reason to launch another adaptive training run.

\section{How Much Is Adaptation Worth?}
\label{sec:headroom}

Finding a bad learned clock and a better fixed clock still does not establish that a more sophisticated adaptive controller is worth building. We therefore ask a separate question: after selecting a strong fixed timing policy, how much additional value remains available to episode-specific timing at all?

\subsection{Random, fixed, and oracle timing}
Let $a_k$ denote micro token accuracy of fixed schedule $[1,k]$. A uniform-random clock chooses one of the six schedules independently of state, giving expected accuracy
\begin{equation}
A_{\mathrm{rand}}=\tfrac16\sum_{k=2}^{7}a_k=0.517907.
\end{equation}
The best single fixed schedule is $[1,5]$ with $A_{\mathrm{fixed}}=0.521206$. For each episode $i$, we then allow an oracle to inspect all six held-out outcomes and choose the schedule with highest episode accuracy. Pooling the selected outcomes gives $A_{\mathrm{oracle}}=0.522557$. This oracle is deliberately unreachable: it uses held-out labels after all counterfactual trajectories have been evaluated.

The remaining headroom above the best fixed clock is only
\begin{equation}
H=A_{\mathrm{oracle}}-A_{\mathrm{fixed}}=0.001351,
\end{equation}
with paired-bootstrap 95\% CI $[0.000858,0.001943]$. Relative to random timing, the best fixed clock captures
\begin{equation}
\frac{A_{\mathrm{fixed}}-A_{\mathrm{rand}}}{A_{\mathrm{oracle}}-A_{\mathrm{rand}}}\approx 0.71
\end{equation}
of the measurable timing value. Timing matters globally, but after choosing a good clock, at most about 29\% of this already-small interval remains for perfect episode-specific selection over the same action space.

\subsection{The residual opportunity is sparse}
The oracle gain is highly concentrated. Only 281 of 3,686 episodes (7.6\%) obtain any additional correct tokens over fixed $[1,5]$, and eight episodes (0.2\%) contribute half of the total micro headroom. Ties are pervasive: 3,252 episodes (88.2\%) share their maximum across at least two schedules and 1,980 (53.7\%) tie across all six. Tie-breaking therefore changes which schedule is nominally credited per episode but not the oracle ceiling itself.

\begin{figure*}[t]
\centering
\begin{subfigure}[t]{0.44\textwidth}
\centering\includegraphics[width=\linewidth]{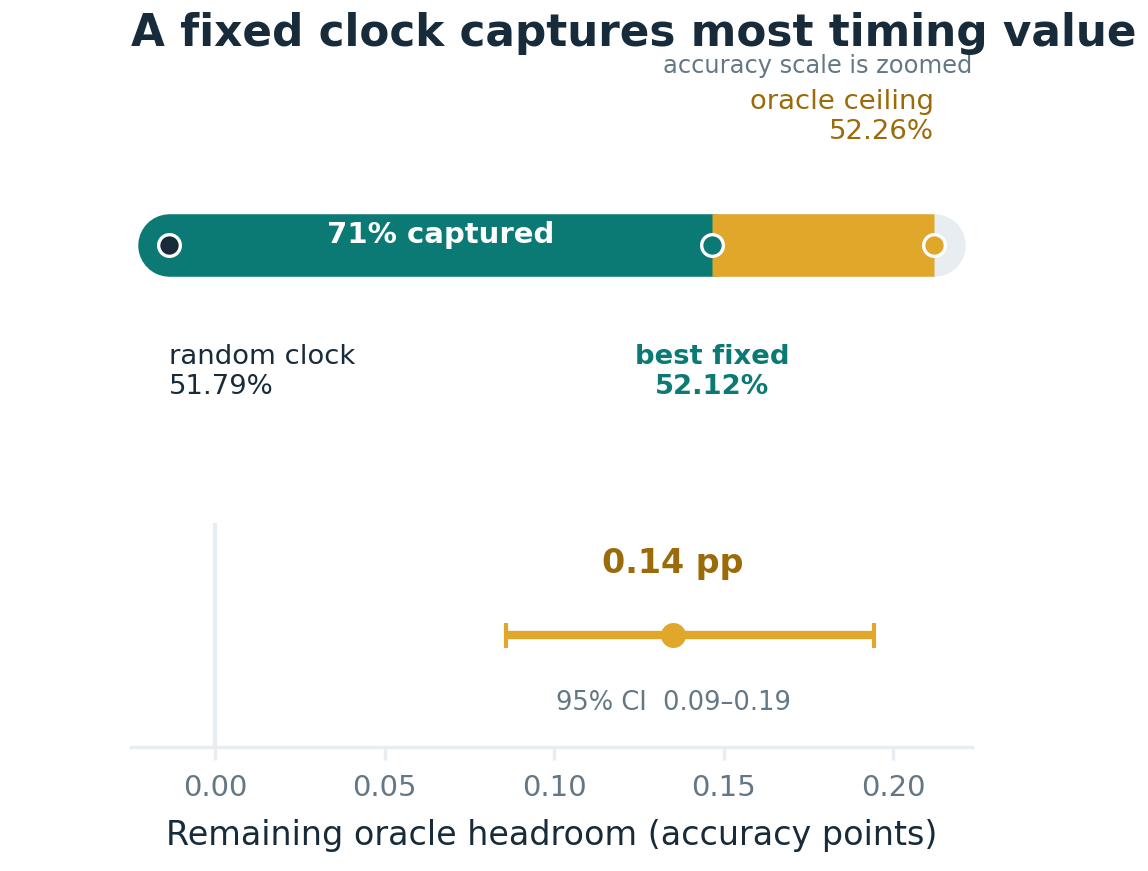}
\caption{Best fixed captures most timing value.}
\end{subfigure}\hfill
\begin{subfigure}[t]{0.44\textwidth}
\centering\includegraphics[width=\linewidth]{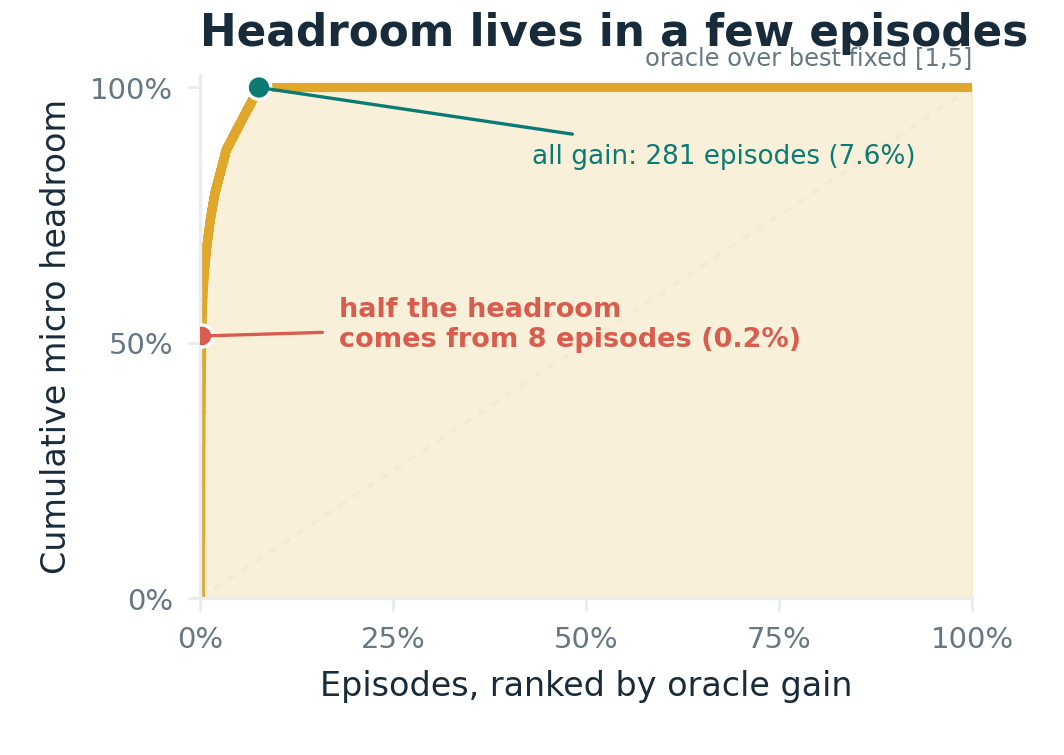}
\caption{Residual headroom is concentrated.}
\end{subfigure}
\caption{Decision-value decomposition over the complete $K=2$ clock space. Left: the best fixed clock captures about 71\% of the improvement from uniform-random timing to the held-out-label oracle. Right: only a small fraction of episodes contributes the remaining gain, and half of that gain comes from eight episodes.}
\label{fig:headroom}
\end{figure*}

\subsection{Adaptation has two prerequisites}
These measurements separate two prerequisites for adaptive supervisory control. First, there must be \emph{heterogeneity of decision value}: different actions must change outcomes. Our frozen interventions clearly satisfy this condition. Second, substantial heterogeneity must remain \emph{after conditioning on a strong non-adaptive baseline}. That condition is much weaker here. Once $[1,5]$ is selected globally, the held-out-label oracle has only a narrow margin left to exploit, and Section~\ref{sec:audit} finds limited evidence that the available state features can identify the rare residual cases.

This changes the design problem from ``learn a timing decision independently on every episode'' to
\begin{equation}
\underbrace{\text{choose a strong global timing prior}}_{\text{most measured value}}
\quad+\quad
\underbrace{\text{learn rare deviations}}_{\text{small residual opportunity}}.
\end{equation}
The oracle is an upper bound, not an achievable target. In this checkpoint and decision space, little episode-specific value remains after selecting a strong fixed clock. This suggests a simple experimental order: measure a strong fixed rule, bound instance-specific value, and only then ask whether state predicts the residual. That separates learning the wrong timing policy from there being little adaptive value left to learn.

\section{Discussion: From Internal Revision to Meta-Level Control}

Our testbed is an internal manager, not an autonomous agent that rewrites another agent or harness; the transferable object is the evaluation problem. Meta-level systems such as GEPA, AlphaEvolve, and Meta-Harness likewise observe evidence, assign credit to revisions, and turn scores into actions \citep{agrawal2025gepa,alphaevolve2025,lee2026metaharness}. Three questions should remain separate: \emph{Does the score vary with state? Does that variation change decisions? Do those changes add outcome value?} Our seeds answer these differently: one becomes a clock and another is behaviorally diverse, yet none beats its strongest forced timing control.

The lesson is not ``use clocks.'' It is to measure adaptive value before treating state dependence or policy diversity as progress. Frozen interventions, counterfactual branching, and strong non-adaptive baselines provide this discipline even when exhaustive enumeration is impossible.

\section{Limitations}

The adaptive audit covers three precommitted seeds, but the exhaustive sweep, oracle analysis, and counterfactual diagnostic are seed-0 analyses; broader tasks, architectures, and compute regimes remain untested. ConceptARC-mini is a mechanism-oriented setting. The original fixed $P=4/P=6$ seed-0 runs did not persist final weights, so those matched-controller checkpoints are deterministic reconstructions whose training summaries reproduce the recorded values to reported precision. The $K=2$ sweep spans six schedules on one anti-diagonal of dwell pairs: it establishes order sensitivity, not independent dwell effects. The held-out-label oracle is an unattainable upper bound over these clocks. Micro and episode-averaged accuracy are labeled separately, and residual trigger variation is not interpreted causally.

\section*{Responsible-Use Statement}

This work highlights a risk of \emph{false assurance}: supervisory scores or diverse actions can appear adaptive without improving downstream outcomes. We recommend strong matched-compute baselines and frozen interventions or counterfactual rollouts to test realized decisions. This diagnostic testbed does not imply that fixed supervision is generally preferable.

\appendix
\section{Additional Evaluation Details}

\paragraph{Evidence hierarchy.}
Headline claims use the three precommitted 400-step adaptive audits and their within-checkpoint forced-schedule comparisons. The exhaustive $K=2$ sweep, oracle analysis, and counterfactual persistence diagnostic are deeper mechanistic analyses of seed 0. The corrected three-seed fixed-period matrix is contextual control evidence. The 96-step adaptive experiment is used only as mechanism sanity evidence; counterfactual persistence v2 stopped before policy training.

\paragraph{Schedule notation.}
An emission after pass $m$ changes the commitment first consumed on pass $m+1$. For $M=8$, $[1,k]$ therefore contains the mandatory initial emission plus one replacement. The corresponding two commitment dwells sum to seven worker passes. This convention explains why reversed pairs such as $[1,2]$ and $[1,7]$ exchange the two dwell lengths.

\paragraph{Calibration.}
Threshold calibration optimizes only the mean intervention count on the 921-episode calibration split. The selected threshold is then frozen and evaluated once on the disjoint 3,686-episode final split. This procedure prevents task labels or final accuracy from selecting the operating point.

\paragraph{Bootstrap.}
Schedule comparisons and oracle headroom use paired episode resampling with 10,000 bootstrap draws. Pairing preserves the fact that all six clocks are evaluated on identical episodes. Confidence intervals reported in the main text are percentile intervals.

\clearpage
\section{Supplementary Quantitative Results}

\begin{table}[h]
\centering
\caption{Three-seed adaptive audit and frozen-checkpoint forced controls.}
\small
\begin{tabular}{@{}cccccc@{}}
\toprule
Seed & Final $K$ & Adaptive acc. & Best forced & Best acc. & Gap (pp) \\
\midrule
0 & 2.0008 & 0.467015 & $[1,4]$ & 0.487511 & $-2.0496$ \\
1 & 1.9674 & 0.486149 & $[1,6]$ & 0.489535 & $-0.3386$ \\
2 & 1.9135 & 0.474947 & $[1,2]$ & 0.475515 & $-0.0568$ \\
\bottomrule
\end{tabular}
\end{table}

\begin{table}[h]
\centering
\caption{Headline quantitative checks. Metrics are not mixed across aggregation families.}
\small
\begin{tabular}{@{}lll@{}}
\toprule
Experiment & Quantity & Result \\
\midrule
Adaptive seed 0 & final mean $K$ & 2.0008 \\
Adaptive seed 0 & episodes selecting $[1,2]$ & 99.19\% \\
Adaptive seed 0 & position-explained $\beta$ variance & 96.76\% \\
Matched training & adaptive ep.-avg. token acc. & 0.4670 \\
Matched training & fixed $P=4$ / $P=6$ & 0.4813 / 0.4856 \\
Frozen adaptive weights & forced $[1,2]$ & 0.4670 \\
Frozen adaptive weights & forced $[1,4]$ / $[1,6]$ & 0.4875 / 0.4866 \\
$K=2$ sweep & best fixed $[1,5]$ micro acc. & 0.52121 \\
$K=2$ sweep & uniform-random micro acc. & 0.51791 \\
$K=2$ sweep & oracle micro acc. & 0.52256 \\
$K=2$ sweep & oracle headroom over best fixed & 0.00135 \\
\bottomrule
\end{tabular}
\end{table}

\begin{table}[h]
\centering
\caption{Preregistered reversed-order comparisons, reported as longer-first minus shorter-first micro token accuracy.}
\small
\begin{tabular}{@{}lcc@{}}
\toprule
Comparison & Difference & 95\% paired bootstrap CI \\
\midrule
$[1,7]-[1,2]$ & 0.00924 & [0.00847, 0.01006] \\
$[1,6]-[1,3]$ & 0.00405 & [0.00346, 0.00472] \\
$[1,5]-[1,4]$ & 0.00097 & [0.00075, 0.00123] \\
\bottomrule
\end{tabular}
\end{table}

\section{Reproducibility Notes}
The primary model uses hidden and goal dimensions 32, two attention heads, one high-level and one low-level layer, and one cycle at each level. The primary objective combines Stablemax token loss, halt/continue losses, and directional loss with $\lambda_F=0.05$ and intervention weight $\eta=0$. Training uses AdamATan2, batch size 4, learning rate $10^{-4}$, puzzle-embedding learning rate $10^{-2}$, weight decay 0.1, no warmup, 400 optimizer steps, and CPU execution. The adaptive replication uses seeds 0--2; the contextual fixed-period matrix also uses seeds 0--2.

Each adaptive step-400 checkpoint was frozen before its schedule-control evaluation. SHA-256 digests are:
\begin{flushleft}\footnotesize\ttfamily
seed 0: 165586ca87a5ca473b342ccb4342afd72890817498c87319964356aa37313889\\
seed 1: 688c870db32b2ccb0209f9eb93395e5bcf9cb886963fea5055b81284d627d8e8\\
seed 2: 67526940b33f324fafe17565cc31f231bae01a84fc1adcb2710e2e254819d2a5
\end{flushleft}
Fixed-compute completion was asserted at runtime and terminal goal emissions were suppressed. Counterfactual features and targets were detached, and matched branches were simulated from identical latent carries. The implementation uses numerically safe cosine normalization and bounded gate/halting features. The intervention-loss normalization assumes the fixed $M=8$ regime and should be changed before varying the compute budget.


\begin{thebibliography}{10}

\bibitem[Agrawal et~al.(2025)Agrawal, Tan, Soylu, et~al.]{agrawal2025gepa}
Lakshya~A. Agrawal, Shangyin Tan, Dilara Soylu, et~al.
\newblock GEPA: Reflective prompt evolution can outperform reinforcement learning.
\newblock \emph{arXiv preprint arXiv:2507.19457}, 2025.

\bibitem[AlphaEvolve Team(2025)]{alphaevolve2025}
AlphaEvolve Team.
\newblock AlphaEvolve: A Gemini-powered coding agent for designing advanced algorithms.
\newblock Google DeepMind, 2025.

\bibitem[Bacon et~al.(2017)Bacon, Harb, and Precup]{bacon2017optioncritic}
Pierre-Luc Bacon, Jean Harb, and Doina Precup.
\newblock The option-critic architecture.
\newblock In \emph{AAAI}, 2017.

\bibitem[Banino et~al.(2021)Banino, Balaguer, and Blundell]{banino2021pondernet}
Andrea Banino, Jan Balaguer, and Charles Blundell.
\newblock PonderNet: Learning to ponder.
\newblock In \emph{ICML Deep Learning Workshop}, 2021.

\bibitem[Campos et~al.(2018)Campos, Jou, Gir\'o-i-Nieto, Torres, and Chang]{campos2018skiprnn}
V\'ictor Campos, Brendan Jou, Xavier Gir\'o-i-Nieto, Jordi Torres, and Shih-Fu Chang.
\newblock Skip RNN: Learning to skip state updates in recurrent neural networks.
\newblock In \emph{ICLR}, 2018.

\bibitem[Graves(2016)]{graves2016act}
Alex Graves.
\newblock Adaptive computation time for recurrent neural networks.
\newblock \emph{arXiv preprint arXiv:1603.08983}, 2016.

\bibitem[Lee et~al.(2026)Lee, Nair, Zhang, Lee, Khattab, and Finn]{lee2026metaharness}
Yoonho Lee, Roshen Nair, Qizheng Zhang, Kangwook Lee, Omar Khattab, and Chelsea Finn.
\newblock Meta-Harness: End-to-end optimization of model harnesses.
\newblock \emph{arXiv preprint arXiv:2603.28052}, 2026.

\bibitem[Sutton et~al.(1999)Sutton, Precup, and Singh]{sutton1999options}
Richard~S. Sutton, Doina Precup, and Satinder Singh.
\newblock Between MDPs and semi-MDPs: A framework for temporal abstraction in reinforcement learning.
\newblock \emph{Artificial Intelligence}, 112(1--2):181--211, 1999.

\bibitem[Vezhnevets et~al.(2017)Vezhnevets, Osindero, Schaul, et~al.]{vezhnevets2017fun}
Alexander~S. Vezhnevets, Simon Osindero, Tom Schaul, et~al.
\newblock FeUdal Networks for hierarchical reinforcement learning.
\newblock In \emph{ICML}, 2017.

\bibitem[Wang et~al.(2025)Wang, Li, Sun, et~al.]{wang2025hrm}
Guan Wang, Jin Li, Yuhao Sun, et~al.
\newblock Hierarchical Reasoning Model.
\newblock \emph{arXiv preprint arXiv:2506.21734}, 2025.

\end{thebibliography}
\end{document}